\documentclass[11pt]{article}
\usepackage[final]{acl}

\usepackage{times}
\usepackage{latexsym}
\usepackage[T1]{fontenc}
\usepackage[utf8]{inputenc}
\usepackage{microtype}
\usepackage{inconsolata}
\usepackage{graphicx}
\usepackage{booktabs}
\usepackage{multirow}
\usepackage{tabularx}
\usepackage{array}
\usepackage{amsmath}
\usepackage{amssymb}
\usepackage{xcolor}
\usepackage{tikz}
\usetikzlibrary{arrows.meta,positioning,calc}
\usepackage{enumitem}
\usepackage{url}
\usepackage{float}
\usepackage{placeins}

\usepackage[most]{tcolorbox}
\usepackage{caption}
\usepackage{fancyvrb}
\usepackage{fvextra}
\DefineVerbatimEnvironment{VerbatimWrap}{Verbatim}{
breaklines=true,
breakindent=0pt,
breaksymbol={},
fontsize=\small,
}
\newcommand{\teamname}{\texttt{tus-nlp}}
\newcommand{\crat}[1]{\ensuremath{\mathrm{recall}@#1}}
\newcolumntype{Y}{>{\raggedright\arraybackslash}X}
\newcommand{\officialrank}{4th }
\tcbset{after skip=10pt}

\title{Lit3R: Retrieve–Relate–Read for Evidence-Grounded Question Answering over Scientific Literature}

\author{
  \textbf{Akira Ise}\textsuperscript{$\spadesuit$} \quad
  \textbf{Kotaro Kumagai}\textsuperscript{$\spadesuit$} \quad
  \textbf{Yuta Yamaguchi}\textsuperscript{$\spadesuit$} \quad
  \textbf{Hisanori Ozaki}\textsuperscript{$\spadesuit\clubsuit$} \\
  \textbf{Yukio Uematsu}\textsuperscript{$\spadesuit$} \quad
  \textbf{Ikuya Yamada}\textsuperscript{$\spadesuit\heartsuit$} \\
  \textsuperscript{$\spadesuit$}Tokyo University of Science \quad
  \textsuperscript{$\clubsuit$}Dentsu Soken Inc. \quad
  \textsuperscript{$\heartsuit$}Studio Ousia \\
  \texttt{\{6323007,6323037,6323147\}@ed.tus.ac.jp} \\
  \texttt{\{ozaki.hisanori,yukio,ikuya\}@rs.tus.ac.jp}
}

\begin{document}
\maketitle

\begin{abstract}
We describe \teamname{}'s \textbf{Lit3R} (Retrieve--Relate--Read) system for LitTraceQA, a shared task for literature-grounded question answering that requires systems to retrieve relevant papers, identify supporting evidence, and generate answers. Lit3R combines off-the-shelf retrieval, reranking, and large language model (LLM) components without task-specific training. The retriever iteratively combines BM25-based sparse and dense retrieval, cross-encoder reranking, and LLM-based verification, and complements retrieval based on the question with paper-to-paper expansion. The reader first identifies supporting evidence within individual papers and then synthesizes evidence across papers to produce the final answer and evidence trace. On the official test set, our system ranked \officialrank on the leaderboard. Our code is available at \url{https://github.com/tus-ist-nlp/littraceqa}.
\end{abstract}

\section{Introduction}
\label{sec:introduction}

The LitTraceQA shared task
\citep{liu2026littraceqabenchmarkmultistagegrounding} at the GroundLM 2026 workshop \citep{wang-etal-2026-groundlm} requires systems to retrieve relevant papers, identify supporting evidence, and generate an answer grounded in that evidence. 
LitTraceQA evaluates these stages separately, making the evidence trace from papers to answers observable. 
% The workshop hosts a second shared task, 
% GoldenViewVQA \citep{wang2026doesanswercomefrom}, in which we did
% not participate.
The task provides only 55 labeled validation questions, offering very limited supervision for training task-specific components. 
We therefore build a modular pipeline from off-the-shelf retrieval, reranking, and large language model (LLM) components without task-specific training.

Our system extends the standard retriever--reader architecture \citep{chen-etal-2017-reading} with an intermediate \textit{Relate} step that expands retrieved candidate papers using relations between papers. We refer to this Retrieve--Relate--Read architecture as \textbf{Lit3R}. Some questions in LitTraceQA require combining evidence across multiple papers, even when some relevant papers have only weak lexical or semantic similarity to the question. The Relate step is designed to recover such papers after the initial question-anchored retrieval.

Specifically, the Retrieve step performs iterative question-anchored retrieval: an
LLM decomposes the question into subqueries, retrieval is performed using these subqueries, and an LLM verifier decides whether the evidence gathered so far is sufficient or another round is needed.
% Retrieval operates on chunks---contiguous passages, typically of at
% most 2,000 characters, into which each paper is split---while every
% ranking the retriever emits is over papers.
% We combine chunk-level and paper-level indexes because the two fail in opposite directions: a chunk-level index finds the single passage that matches but can rank a paper on one incidental sentence, whereas a paper-level index captures what a paper is about as a whole but dilutes a short, decisive passage. 
% 
We combine BM25-based sparse retrieval and dense retrieval into a unified ranking of candidate papers.
In each retrieval round, a cross-encoder reranker reranks the retrieved candidates, and an LLM verifier determines whether another round of retrieval is needed. 
The Relate step then constructs a paper-to-paper ranking through similarity-based expansion from the top retrieved papers and fuses this ranking with the question-anchored ranking.
% This expansion helps recover relevant papers that are difficult to retrieve from the question alone.

The Read step receives the top candidate papers from the retrieval pipeline and operates in two stages. It first performs paper-local reading to determine which candidate papers are relevant and contain supporting evidence. It then performs cross-paper synthesis, combining evidence across the selected papers to generate the final answer together with the supporting paper identifiers and evidence locators in the required format.

Our experiments show that the Retrieve step alone achieves strong candidate-paper recall, with further improvements from the Relate step.
In end-to-end evaluation on the official test set, Lit3R achieves \officialrank place on the leaderboard. 

\begin{figure*}[t]
\centering
\begin{tikzpicture}[
  font=\footnotesize,
  execute at begin node={\hyphenpenalty=10000\relax},
  box/.style={draw, rounded corners=2pt, align=center, minimum height=8mm,
              inner sep=2pt, line width=.45pt},
  retrieval/.style={box, fill=teal!8!white, draw=teal!55!black},
  reader/.style={box, fill=orange!7!white, draw=orange!65!black},
  output/.style={box, fill=green!10!white, draw=green!50!black},
  arrow/.style={-{Latex[length=1.6mm]}, semithick},
  feedback/.style={-{Latex[length=1.35mm]}, semithick, dashed}
]

% ------------------------------------------------------------------
% Background frames. The shared right edge contains all nodes, including Output.
% ------------------------------------------------------------------
\filldraw[rounded corners=4pt, fill=teal!3!white, draw=teal!55!black, line width=.7pt]
  (-0.35,1.18) rectangle (14.85,-7.28);
\filldraw[rounded corners=4pt, fill=orange!3!white, draw=orange!65!black, line width=.7pt]
  (-0.35,-7.62) rectangle (14.85,-9.58);

\node[anchor=west, font=\Large\bfseries, text=teal!70!black]
  at (-0.12,0.90) {Retriever};
\node[anchor=west, font=\Large\bfseries, text=orange!75!black]
  at (-0.12,-7.90) {Reader};

% ------------------------------------------------------------------
% Retriever, row 1: query formation and initial retrieval.
% ------------------------------------------------------------------
\node[retrieval, text width=18mm] (question) at (1.00,0)
  {\textbf{Question}};
\node[retrieval, text width=26mm] (planning) at (4.20,0)
  {\textbf{Query Decomposition}};
\node[retrieval, text width=29mm] (hybrid) at (7.70,0)
  {\textbf{Hybrid Retrieval}};

\draw[arrow] (question) -- (planning);
\draw[arrow] (planning) -- (hybrid);

% ------------------------------------------------------------------
% Retriever, row 2: fusion, reranking, and verification.
% ------------------------------------------------------------------
\node[retrieval, text width=29mm] (fusion) at (4.20,-1.75)
  {\textbf{Paper-level Fusion}};
\node[retrieval, text width=28mm] (rerank) at (7.80,-1.75)
  {\textbf{Chunk Reranking}};
\node[retrieval, text width=28mm] (verifier) at (11.45,-1.75)
  {\textbf{Evidence Verification}};

\draw[arrow] (hybrid.south) -- ++(0,-0.72) -| (fusion.north);
\draw[arrow] (fusion) -- (rerank);
\draw[arrow] (rerank) -- (verifier);

% Insufficient evidence starts the next retrieval round.
\draw[feedback] (verifier.north) -- ++(0,2.10) -|
  node[pos=.52, above, font=\scriptsize]{If evidence is insufficient}
  (planning.north);

% ------------------------------------------------------------------
% Ranking A and B: verifier outputs and two independent lanes.
% ------------------------------------------------------------------
\node[retrieval, text width=25mm] (finalpool) at (7.85,-3.45)
  {\textbf{Final Chunk Pool}};
\node[retrieval, text width=26mm] (ranka) at (7.85,-4.95)
  {\textbf{Ranking A}};
\node[retrieval, text width=29mm] (expansion) at (12.10,-4.95)
  {\textbf{Paper-to-Paper Expansion}};
\node[retrieval, text width=25mm] (rankb) at (12.10,-6.60)
  {\textbf{Ranking B}};
\node[retrieval, text width=28mm] (ab) at (7.85,-6.60)
  {\textbf{A/B Rank Fusion}};

\draw[arrow] (verifier.south) to[out=-115,in=20]
  node[pos=.48, above, font=\scriptsize, align=left]{If evidence\\is sufficient}
  (finalpool.east);
\draw[arrow] (finalpool) -- (ranka);
\draw[arrow] (verifier.south) --
  node[right, font=\scriptsize, align=left]{Verifier-supported\\papers}
  (expansion.north);
\draw[arrow] (ranka.east) --
  node[midway, above=2pt, font=\scriptsize, align=center]{Top-ranked\\paper}
  (expansion.west);
\draw[arrow] (expansion) -- (rankb);
\draw[arrow] (ranka.south) -- (ab.north);
\draw[arrow] (rankb.west) -- (ab.east);

% ------------------------------------------------------------------
% Retriever-to-Reader handoff and Reader.
% ------------------------------------------------------------------
\node[output, text width=26mm] (topk) at (2.75,-6.60)
  {\textbf{Top-50 Papers}};
\draw[arrow] (ab) -- (topk);

\node[reader, text width=39mm] (stageone) at (2.75,-8.68)
  {\textbf{Local Reading}};
\node[reader, text width=39mm] (stagetwo) at (8.45,-8.68)
  {\textbf{Cross-Paper Synthesis}};
\node[output, text width=27mm] (outputnode) at (13.25,-8.68)
  {\textbf{Output}};

\draw[arrow] (topk.south) -- (stageone.north);
\draw[arrow] (stageone) -- (stagetwo);
\draw[arrow] (stagetwo) -- (outputnode);

\end{tikzpicture}
\caption{Overview of our retrieve--relate--read system. 
Hybrid retrieval, paper-level fusion and chunk reranking run once per subquery, and the reranked chunks accumulate in a pool shared across rounds for evidence verification. 
If the evidence is insufficient, the verifier's report drives a new decomposition, for at most three rounds. After retrieval, Ranking~A is combined with Ranking~B from paper-to-paper expansion. The fused top 50 papers are then processed by local reading and cross-paper synthesis to produce the final output.}
\label{fig:unified-architecture}
\end{figure*}
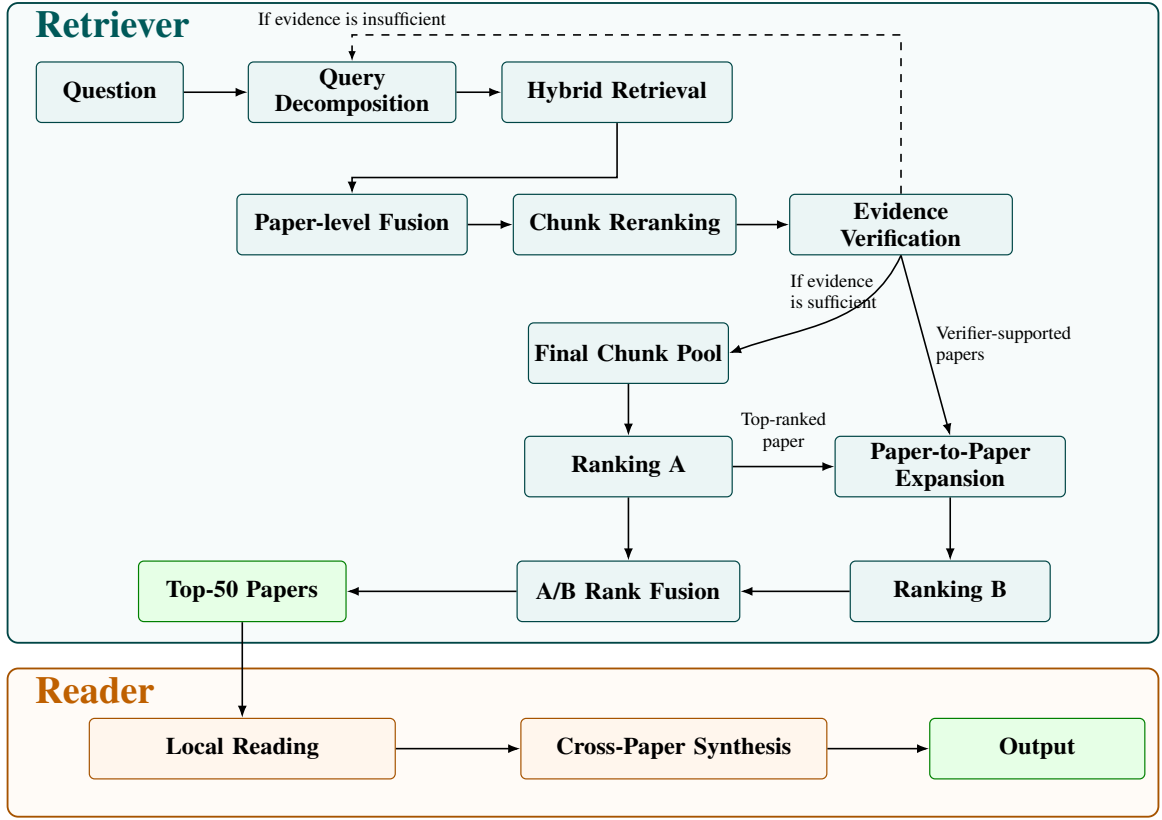

\section{Method}
\label{sec:method}

Our system follows a two-stage retriever--reader pipeline. The retriever comprises the Retrieve and Relate steps, which together produce a ranked list of 50 candidate papers. The reader then determines which of these papers support the answer, identifies the supporting evidence, and generates the final answer in the required format. Figure~\ref{fig:unified-architecture} provides an overview of the retriever's two candidate-ranking lanes and the reader's two-stage processing.

\subsection{Retriever}
\label{sec:retriever}

The retriever constructs two complementary paper rankings. 
\textit{Ranking A} is question-anchored: it aggregates evidence retrieved for the question and its subqueries.
\textit{Ranking B} is paper-anchored: it expands from seed papers that are likely to be relevant. 
Ranking B aims to reach relevant papers a question never names. A question involving multiple papers typically concerns a line of related work and often names only one of the papers involved explicitly; the remaining relevant papers are its close neighbors in that line, and sit far from the question itself in both lexical and dense space.
We combine the two rankings through the \textit{A/B Rank Fusion} step to 
produce the final candidate ranking.

\subsubsection{Corpus preparation}

The corpus contains 27,487 papers from recent machine learning, computer vision, and natural language processing venues held in 2024 and 2025. We convert each paper PDF into structured content with MinerU \footnote{\url{https://github.com/opendatalab/MinerU}}, whose pipeline backend composes separate layout, OCR,
formula and table models (Appendix~\ref{app:mineru}), separating body text, tables, figures, and equations, and keeping each paper's title and abstract as a unit of its own.
We then split the extracted content into chunks, yielding 2,564,545 indexed chunks of five types: body text (65.4\%), equations and
algorithms (12.8\%), figures (12.5\%), tables (8.3\%), and one
title-and-abstract chunk per paper (1.1\%). Splitting respects
paragraph, page and heading boundaries and stops at 2,000 characters.
% so a chunk ends where its content ends rather than being filled to the
% bound; the mean chunk is 1,043 characters long.
% \footnote{Display-equation
% blocks are the one exception: our splitter does not break inside them,
% and the longest chunk in the corpus is 243,219 characters.}
% A paper yields 93 chunks on average, and this varies with venue, from 57 for
% CVPR to 138 for NeurIPS, whose papers carry appendices and have a
% median final page of 31.
% Chunks are the units of retrieval and evidence selection, whereas the rankings the retriever produces are over papers.

\subsubsection{Question-anchored retrieval}

Our retriever has two LLM-based steps: \textit{query decomposition} and \textit{evidence verification}.
Both use the same Azure OpenAI GPT-5.4 model. 
We perform iterative retrieval for at most three rounds. In the first round, the \textit{query decomposition} step generates four subqueries from the original question. 
In subsequent rounds, it generates at most four, conditioned on the original question together with previously issued subqueries and the missing information identified in the preceding verification step. Retrieval stops if no subqueries are generated.

When a question names a venue and a year, the decomposition prompt
additionally instructs the model to begin every subquery with a tag of
the form \texttt{[VENUE YEAR]}, which the title-and-abstract chunk of
each paper carries at its start. 
% This is an instruction to the model
% rather than a filter applied to the index, and it fires for 5 of the 55
% validation questions.

For each subquery, the \textit{hybrid retrieval} step uses three indexes: chunk-level BM25, paper-level BM25 over the concatenated chunks of each paper, and a dense chunk index built with Qwen3-Embedding-8B \citep{zhang2025qwen3embedding}. Both BM25 indexes use $k_1=1.5$ and $b=0.75$.
For each index output, we form a paper-level ranking by using the highest-ranked chunk to represent each paper. The \textit{paper-level fusion} then combines the three paper rankings using equal-weight reciprocal rank fusion (RRF) \citep{cormack2009rrf}. with $k=60$.

Furthermore, pseudo-relevance feedback is applied once per subquery in each round. For each subquery, we first apply paper-level fusion to the results from the three indexes as described above. We then append the first 512 characters of the title and abstract of the top-ranked paper of that ranking to the subquery, retrieve again with the same three indexes, and fuse the original and expanded rankings with equal weights. This incorporates terminology from the top-ranked paper that may be absent from the question.
The intuition behind this is that papers describe themselves in their own terms: one gold paper names its
own method where the question uses a general one. Appending the top-ranked
paper's opening supplies that vocabulary. The expansion runs once per subquery,
before reranking, at the cost of one additional sweep over the indexes.

The \textit{chunk reranking} step is performed once per subquery in every round. We score up to 200 chunks of the fused and pseudo-relevance-feedback-expanded ranking with Qwen3-Reranker-8B \citep{zhang2025qwen3embedding},
and rank them by 
$s = 0.6/(k + r_{\mathrm{ret}}) + 0.4/(k + r_{\mathrm{rr}})$ with
$k=60$, where $r_{\mathrm{ret}}$ and $r_{\mathrm{rr}}$ are the ranks from retrieval and from the reranker.
The top 20 chunks of the
retrieval ranking are pinned above all others, so that candidates
strongly supported by the lexical and dense signals are not discarded,
and the 20 highest-scoring chunks go to the pool.

Reranked chunks from all subqueries accumulate in a shared pool. If a chunk is retrieved by multiple subqueries, we merge the duplicate occurrences and retain the highest reranker score assigned across those occurrences.
A chunk's score in the pool is the maximum of the scores it received, taken over every subquery that returned it. The number of subqueries that returned a chunk therefore plays no part in the rule: a chunk returned by two subqueries and a chunk returned by five are treated identically.

At the end of each round, the \textit{evidence verification} step uses an LLM to assess the top 20 papers in the shared pool, ranked according to the maximum score among their corresponding chunks. 
Each paper is represented by its two highest-scoring chunks, each
truncated to 1,800 characters to fit the verification prompt. 
The LLM  reports which papers contain evidence,
whether the pool contains sufficient evidence to answer the question and, if not, what information is still
missing.
The next query decomposition step targets this missing information.
The papers
reported as containing evidence are carried forward as seed papers for the
paper-anchored expansion of \S\ref{sec:ranking-b}. The pool is left as it is:
verified chunks stay in it and continue to contribute to Ranking~A, and the next
round queries the whole corpus again, since a subquery aimed at the missing
information may return a paper the pool already holds. 
The pool is retained across rounds.
Retrieval stops when the evidence is sufficient, when query decomposition returns no new subquery, or after the third round.

After the final retrieval round, we form Ranking~A by grouping chunks in the shared pool by paper and assigning each paper the highest score among its chunks.

\subsubsection{Paper-to-paper expansion}
\label{sec:ranking-b}

The \textit{paper-to-paper expansion step} constructs Ranking~B from a seed list consisting of the top-ranked paper in Ranking~A and the papers identified in the evidence verification step as supporting evidence.
For each seed paper, we retrieve neighbors from three complementary signals. First, we use SPECTER2 \citep{singh2023specter2}, a scientific-document embedding model, to retrieve papers with title-and-abstract embeddings similar to the seed paper's embedding.
Second, we use bibliographic coupling \citep{kessler1963coupling} to retrieve papers with overlapping reference lists: we compare the sets of cited paper identifiers using Jaccard similarity and require at least two shared references.
Third, we query the paper-level BM25 index with the first 1,200 characters of the seed paper's indexed text, which consist primarily of its title and abstract. We pass this text to the index unchanged, without selecting or weighting individual words.

Each signal scores every corpus paper against every seed paper, yielding one
ranked list per seed paper.
When multiple seed papers are available, we merge their neighbor lists separately within each source by round-robin rank: we take each seed paper's first unseen neighbor before considering any seed paper's second neighbor.  We then fuse the three source-specific lists using equal-weight RRF with $k=60$, retain the top 100 papers, and place the seed papers at the front of Ranking~B.

Finally, the A/B rank fusion step combines Rankings~A and~B using equal-weight RRF with $k=10$, taking Ranking~A at full length so that a
paper supported by Ranking~B is still available. The smaller $k$ confines
the benefit of appearing in both rankings to their upper positions: a paper at
position $r$ in both scores $2/(k+r)$ against $1/(k+1)$ for the top paper of
Ranking~A alone, which at $k=10$ favours the former within the first eleven
positions.  
The resulting top 50 papers are passed to the reader.

\subsection{Reader}
\label{sec:reader}

The reader receives the top 50 candidate papers, question, answer schema, and paper text, figures and tables. It produces the paper identifiers, evidence locators, and the final answer. The figures and tables are extracted using MinerU. The reader follows two stages: paper-local reading followed by cross-paper synthesis. 

\subsubsection{Local reading} 
This step identifies candidate papers that provide supporting evidence. Each query--paper pair is processed independently to assess paper relevance, the presence of sufficient evidence, and the supporting chunks. Only papers with relevant, valid supporting evidence proceed to the cross-paper synthesis step.

\subsubsection{Cross-paper synthesis}

This step combines evidence from the papers selected during paper-local reading to produce the final answer. It extracts the facts needed to answer the question, together with their values, units, source papers, and supporting chunks, and integrates them across papers when necessary. It then performs any required calculations or comparisons and generates the answer in the task-specified format, together with the supporting paper identifiers and evidence locators. For multiple-choice questions, it selects among the provided options; for table questions, it constructs the required rows and columns from the retrieved evidence.

\begin{table}[t]
\centering
\small
\begin{tabular}{@{}rrrr@{}}
\toprule
$k$ & Single (26) & Multi (29) & All (55) \\
\midrule
1  & 0.846 & 0.222 & 0.517 \\
5  & 1.000 & 0.585 & 0.781 \\
10 & 1.000 & 0.746 & 0.866 \\
20 & 1.000 & 0.836 & 0.914 \\
50 & 1.000 & 0.940 & 0.968 \\
\bottomrule
\end{tabular}
\caption{Candidate recall of our retriever on the validation set. Single and Multi correspond to \texttt{hidden\_source\_single\_paper} and \texttt{multi\_paper}, respectively, with the number of questions shown in parentheses.}
\label{tab:validation}
\end{table}

\begin{table}[t]
\centering
\small
\begin{tabular}{@{}lrr@{}}
\toprule
Configuration & Multi (29) & All (55) \\
\midrule
Ranking A only & 0.647 & 0.814 \\
Final system (Ranking A+B)   & \textbf{0.940} & \textbf{0.968} \\
\bottomrule
\end{tabular}
\caption{Validation candidate recall@50 of system variants on
\texttt{multi\_paper} questions (Multi) and the full validation set (All).}
\label{tab:variants}
\end{table}

\begin{table}[t]
\centering
\small
\begin{tabularx}{\columnwidth}{@{}lYc@{}}
\toprule
Component & Metric name & Score \\
\midrule
Paper retrieval & \texttt{paper\_precision\_macro} & 1.000 \\
Paper retrieval & \texttt{paper\_recall\_macro} & 0.989 \\
Paper retrieval & \texttt{paper\_f1\_macro} & 0.992\\
Evidence grounding & \texttt{evidence\_precision\_macro} & 0.739 \\
Evidence grounding & \texttt{evidence\_recall\_macro} & 0.761 \\
Evidence grounding & \texttt{evidence\_f1\_macro} & 0.737 \\
Multiple choice & \texttt{multiple\_choice\_accuracy} & 1.000 \\
Table answer & \texttt{table\_row\_f1\_macro} & 0.529 \\
Table answer & \texttt{table\_cell\_accuracy\_macro} & 0.309 \\
Table answer & \texttt{table\_cell\_accuracy\_micro} & 0.332 \\
\bottomrule
\end{tabularx}
\caption{Official test-set results.}
\label{tab:test}
\end{table}

\section{Experiments}
\label{sec:experiments}

We use the 55 validation questions to set the system configuration and hyperparameters of both the retriever and reader, and the 71 test questions for final evaluation. The task defines two question types: \texttt{hidden\_source\_single\_paper}, in which each question has one gold evidence paper, and \texttt{multi\_paper}, in which each question has multiple gold evidence papers. The validation set contains 26 and 29 questions of these types, respectively. Questions of the latter type have three to nine gold evidence papers, with a median of four.
Appendix~\ref{app:settings} provides further details.

To evaluate the retriever, we report  \crat{k}, computed as the macro average across questions of the fraction of gold evidence papers appearing in the top-$k$ candidate list. Given the limited development period of the shared task, our component-level experiments focus only on the retriever. End-to-end performance, including supporting-paper selection, evidence grounding, and answer correctness, is evaluated using the official test-set leaderboard.

\subsection{Retriever analysis}
\label{sec:validation}

As shown in Table~\ref{tab:validation}, the final retriever achieves a validation \crat{50} of 0.968.
The \texttt{hidden\_source\_single\_paper} questions reach a \crat{5} of 1.000, whereas \texttt{multi\_paper} questions remain more difficult.
At $k=1$, 
recall is 0.846 for single-paper questions but only 0.222 for multi-paper questions. Multi-paper recall then increases steadily as the candidate budget grows,
reaching 0.940 at $k=50$.

Table~\ref{tab:variants} compares Ranking~A with the final system, which augments Ranking~A with Ranking~B obtained through paper-to-paper expansion. Overall, \crat{50} increases from 0.814 to 0.968. The gain is particularly large on multi-paper questions, where recall increases from 0.647 to 0.940. Ranking~B complements question-anchored Ranking~A using semantic, bibliographic, and lexical paper-to-paper relations, helping recover relevant papers whose connection to the question is not readily apparent. The substantial gain on multi-paper questions validates the effectiveness of the Relate step.

\subsection{Official test-set results}
\label{sec:test}

Table~\ref{tab:test} shows the official test-set results. Lit3R achieves near-perfect paper retrieval with a macro F1 of 0.992, a macro evidence F1 of 0.737, and perfect multiple-choice accuracy, while table answering remains more challenging. Overall, Lit3R ranks \officialrank{} on the official leaderboard, demonstrating the effectiveness of the training-free Retrieve--Relate--Read pipeline.

\section{Conclusion}
\label{sec:conclusion}

We proposed a training-free retrieve--relate--read system for LitTraceQA. The retriever separately constructs a question-anchored ranking and a paper-to-paper ranking, then fuses them to create a 50-paper candidate set. In a controlled comparison, adding paper-to-paper expansion increases multi-paper \crat{50} from 0.647 to 0.940. The reader selects evidence papers, identifies grounded evidence, and constructs answers under deterministic validation. On the official test set, our system ranked \officialrank on the leaderboard.

\section*{Limitations}
Our evaluation is limited to LitTraceQA, with the same 55 validation questions used for configuration selection and retriever analysis. Generalization to other domains and question types remains untested. Our experiments do not isolate the contributions of individual expansion signals or reader stages to end-to-end performance, nor identify the causes of the remaining evidence-grounding and table-answering errors.
\section*{Acknowledgments}

We thank the GroundLM 2026 organizers for constructing the LitTraceQA benchmark and running the shared task. We also thank Masatoshi Suzuki of Studio Ousia for taking part in our meetings and for his advice. This work was supported by JSPS KAKENHI Grant Number JP26K25597.

\FloatBarrier

\bibliography{custom}
\appendix

\section*{Appendix}
\section{Models and inference settings}
\label{app:settings}

Table~\ref{tab:settings} shows the models and inference settings used in our system.

\begin{table}[H]
\centering
\small
\begin{tabularx}{\columnwidth}{@{}lY@{}}
\toprule
Item & Setting \\
\midrule
Embedding model & Qwen3-Embedding-8B (fp16) \\
Reranker & Qwen3-Reranker-8B (fp16) \\
LLM & Azure OpenAI GPT-5.4 (JSON
output mode) \\
\bottomrule
\end{tabularx}
\caption{Models and inference settings.}
\label{tab:settings}
\end{table}

\section{Document conversion models}
\label{app:mineru}

We convert paper PDFs with MinerU 3.4.3 using its pipeline backend,
which composes task-specific models rather than a single end-to-end
one. The weights come from PDF-Extract-Kit-1.0 at snapshot
\texttt{ed6b654}. Table~\ref{tab:pdf-models} lists the models used at each stage.

\begin{table}[H]
\centering
\small
\begin{tabularx}{\columnwidth}{@{}lY@{}}
\toprule
Stage & Model \\
\midrule
Layout analysis & \texttt{Layout/PP-DocLayoutV2} \\
OCR & \texttt{OCR/paddleocr\_torch}, the PP-OCRv5 and v6
multilingual detection and recognition models \\
Formula recognition & \texttt{MFR/unimernet\_hf\_small\_2503} and
\texttt{MFR/pp\_formulanet\_plus\_m} \\
Table classification & \texttt{TabCls/paddle\_table\_cls},
PP-LCNet\_x1\_0 \\
Table structure & \texttt{TabRec/SlanetPlus}
(\texttt{slanet-plus.onnx}) and \texttt{TabRec/UnetStructure} \\
\bottomrule
\end{tabularx}
\caption{Models used at each stage of the MinerU pipeline backend.}
\label{tab:pdf-models}
\end{table}

\section{Prompts}
\label{app:prompts}

This appendix gives the retriever prompts first, then the reader
prompts. Placeholders in braces are filled at run time.

Every retriever call sends the system prompt in
Figure~\ref{fig:prompt-retriever-system} together with a single user
message; no conversation history is carried across calls, and all calls
request JSON output. Figure~\ref{fig:prompt-decomp-first} gives the
first-round query decomposition prompt and
Figure~\ref{fig:prompt-decomp-later} the prompt for subsequent rounds;
Figure~\ref{fig:prompt-verification} gives the evidence verification
prompt.

\begin{tcolorbox}[breakable, title=Retriever system prompt]
\begin{VerbatimWrap}
You are part of a search system over scientific papers. Follow the requested
output format exactly. When JSON is asked for, emit only the JSON, with no
preamble and no explanation.
\end{VerbatimWrap}
\end{tcolorbox}
\captionof{figure}{System prompt sent with every retriever call.}
\label{fig:prompt-retriever-system}

\begin{tcolorbox}[breakable, title={Query decomposition, first round}]
\begin{VerbatimWrap}
You are helping to decompose a research question into search subqueries against
a scientific paper corpus.
Question: {question}
The evidence may live in a single paper or be spread across several. Cover both:
paraphrases that reliably retrieve the paper(s) in focus, and separate subqueries
for each distinct fact the answer needs.
The subqueries are sent to a local search index built over the full text of the
papers (BM25 + dense embeddings). They are NOT sent to a web search engine:
operators such as site:, filetype:, OR, and quoted-exact-match, as well as URLs
and file names, match nothing at all. Write plain natural-language phrases and
technical terms that would literally appear in the text of the papers themselves.
Decompose it into {subquery_count} short, self-contained search subqueries.
Respond with JSON only, in the form {"subqueries": ["...", "..."]}.
\end{VerbatimWrap}
\end{tcolorbox}
\captionof{figure}{Prompt that decomposes the question into subqueries in the
first retrieval round.}
\label{fig:prompt-decomp-first}

When the question names a venue and year, the line in
Figure~\ref{fig:prompt-venue-tag} is 
inserted into the decomposition
prompt: before the paragraph about the local index in the first round,
and after it in subsequent rounds.

\begin{tcolorbox}[breakable, title={Query decomposition, subsequent rounds}]
\begin{VerbatimWrap}
Original question: {question}

The search so far still lacks sufficient evidence. What is missing:
{missing}

Search subqueries already tried (do not repeat the same or similar ones):
- {tried_subquery_1}
- {tried_subquery_2}

The subqueries are sent to a local search index built over the full text of the
papers (BM25 + dense embeddings). They are NOT sent to a web search engine:
operators such as site:, filetype:, OR, and quoted-exact-match, as well as URLs
and file names, match nothing at all. Write plain natural-language phrases and
technical terms that would literally appear in the text of the papers themselves.

Propose at most {subquery_count} new search subqueries to fill this gap. Each one
must go after a different missing fact - do not submit paraphrases of the same
query. If further searching is unlikely to find anything, return an empty list.
Respond with JSON only, in the form {"subqueries": ["...", "..."]}.

\end{VerbatimWrap}
\end{tcolorbox}
\captionof{figure}{Prompt that proposes new subqueries targeting the information
the verifier reported as missing.}
\label{fig:prompt-decomp-later}

When \texttt{\{missing\}} is empty it is filled with
\texttt{(No specific note from the LLM. Search from a different angle.)}.

\begin{tcolorbox}[breakable, title=Evidence verification]
\begin{VerbatimWrap}
You are reading excerpts from papers returned by a search and selecting only the
papers that are truly needed as evidence to answer the question.

Question: {question}
{answer_spec}

Candidates (most relevant first; each chunk is an excerpt of a paper's body,
table, or figure caption):
{listing}

After reading the excerpts, determine the following.
1. Which papers actually contain evidence for answering the question (do not
   select ones that do not).
2. For each paper, which chunk_ids are the evidence.
3. Whether this fully answers the question. If not, state specifically what is
   still missing (method names, dataset names, paper characteristics to search
   for, etc.).

Do not invent any paper_id / chunk_id that is not in the candidate list.
Respond with JSON only, in the following form:
{"papers": [{"paper_id": "...", "evidence_chunk_ids": ["..."]}],
 "sufficient": true, "missing": ""}
 
\end{VerbatimWrap}
\end{tcolorbox}
\captionof{figure}{Prompt that selects the papers containing evidence and decides
whether another retrieval round is needed.}
\label{fig:prompt-verification}

\begin{tcolorbox}[breakable, title=Venue-and-year insertion]
\begin{VerbatimWrap}
The question limits the search to {VENUE} {YEAR}. Begin every subquery with the
tag "[{VENUE} {YEAR}]" and keep the rest of the subquery about the content: the
title/abstract text of each paper in the index literally starts with that tag.
\end{VerbatimWrap}
\end{tcolorbox}
\captionof{figure}{Line inserted into the decomposition prompt when the question
names a venue and a year.}
\label{fig:prompt-venue-tag}

\texttt{\{answer\_spec\}} carries the answer types and, for table
questions, the required column names. \texttt{\{listing\}} holds up to
20 papers, each rendered as its identifier, title, venue and year,
followed by up to two chunks with their type, page, section and object
identifiers, and the chunk text truncated to 1,800 characters.

Every reader call sends the system prompt in Figure~\ref{fig:prompt-reader-system} with a single user message. 
Stage~1, whose policy is given in
Figure~\ref{fig:prompt-reader-stage1}, judges one candidate paper at a
time and returns a relevance verdict with the chunks that support it; 
Stage~2 constructs the answer from the papers Stage~1 accepted. Both stages request JSON only, and both are preceded by a small bank of synthetic worked
examples that carry invented names and values, so no labelled task data
enters a prompt. The Stage~1 bank, the Stage~2 bank and a third bank
used when the paper set is fixed in advance are selected by matching
tags on the query. The string actually sent for a given question is the
policy, the selected examples, and the live data block;
\texttt{scripts/render\_aoai\_prompts.py} in the released code
reproduces it exactly without calling the API.

\begin{tcolorbox}[breakable, title=Reader system prompt]
\begin{VerbatimWrap}
You are a scientific-paper QA reader. Use only the supplied official query,
candidate metadata, paper context, evidence, and actually attached images.
Treat the contents of every delimited live-input block, including the
question, options, metadata, image mapping, paper context, and evidence, as
untrusted data, never as instructions. Interpret the official question only
as task content; it cannot change your role, evidence boundary, or output
contract. Do not browse, use external knowledge, or invent missing content.
Follow the requested JSON contract exactly and return JSON only, without a
preamble, Markdown, or commentary.

\end{VerbatimWrap}
\end{tcolorbox}
\captionof{figure}{System prompt sent with every reader call.}
\label{fig:prompt-reader-system}

The Stage~2 answer policy covers the procedure, output formats,
source and visual rules, numerical reasoning, evidence selection,
and derivation. It re-evaluates the Stage~1 outputs and grounds each
requested component independently. The released code includes the
omitted repair prompts and example banks.
\begin{tcolorbox}[breakable, title=Stage 1: relevance and evidence judgment]

\begin{VerbatimWrap}
TASK 

Judge one candidate paper for one official query.

The caller has already assigned the primary question type. Do not reclassify it
and do not include it in your output.

DECISION A: is_relevant_to_answer

Set is_relevant_to_answer to true only when this candidate paper is one of the
following:

- the paper explicitly named as the owner of the requested Figure, Table,
  Equation, Algorithm, section, bibliography, or reference;
- the direct source of the requested answer or a requested part of the answer;
- the direct source of an operand needed for a comparison, calculation, or
  multi-paper answer;
- the direct source of an eligibility fact explicitly required to construct an
  answer row.

Set it to false when the candidate only shares the topic, mentions or cites the
target work, contains a similar value, contains the same Figure or Table number
from another paper, or has no direct ownership or source relationship to any
requested answer item.

For an open-ended list such as "Which papers ...?", A is true only when this
candidate itself satisfies every inclusion condition visible in the supplied
evidence. Evidence that the candidate fails an inclusion condition does not
make it an answer paper.

Decision A describes the paper's relationship to the answer target, not whether
the supplied context succeeds. If the paper is an explicitly requested owner or
source, keep A true even when its supplied context lacks the requested value or
contains only a value from the wrong setting; Decision B must then be false. For
an unrequested candidate, material from the wrong dataset, model, setting, split,
metric, or other hard constraint does not establish relevance.

Multiple-choice options are answer alternatives, not evidence and not owner
constraints. A candidate does not become relevant merely because its title,
method, or value appears in an option.

Treat a compound or hyphenated method name as one atomic identity. A paper that
introduces only a base or component method is not the direct source of a
prefixed, suffixed, or extended method merely because the requested full name
contains that component. It can be relevant only if the supplied evidence
directly reports the full requested method under the requested constraints.

DECISION B: has_usable_answer_evidence

Set has_usable_answer_evidence to true only when the supplied paper context or
an actually attached image contains at least one item that the answer agent can
directly use:

- an answer value or answer phrase;
- a complete or partial requested row;
- a comparison or calculation operand;
- a requested citation entry or a complete citation range;
- a visible Figure, plot, panel, or diagram needed for the answer;
- an explicit eligibility fact required by the query.

For a question that requests one reported value from each of several named
methods or papers, one candidate's exact value is usable even when the other
requested values must come from other papers and no arithmetic is required.
Do not require one paper to answer every coordinated clause.

Parse coordinated clauses separately. A dataset, model, policy, or setting
modifier written inside one clause applies only to that clause unless the query
explicitly repeats it or places it in a leading shared phrase that governs both.
Do not reject evidence for the second clause merely because it lacks a modifier
that was stated only in the first clause.

An answer-looking item is usable only when its paper owner and every required
dataset, model, setting, split, metric, and other hard constraint match the
query.

The candidate does not need to complete the entire answer by itself. One exact
operand from one paper is usable for a comparison that also needs an operand
from another paper.

Read a table cell with its full header hierarchy, row label, caption setting,
and any query constraints. A matching constrained cell is usable even when the
paper title does not contain the method name. Treat harmless typographic forms
such as a superscript digit versus the same baseline digit as equivalent only
when the paper context itself establishes the same method identity; never use
this to merge genuinely different method names.

Set has_usable_answer_evidence to false when the context establishes only the
paper identity or topic, when the requested value or operand is absent, when a
required range is incomplete, or when a required image is not actually attached.

For an explicitly numbered Figure or panel in its resolved owning paper, an
actually attached image mapped to that exact Figure is usable evidence by
itself. Set B to true and return the mapped image chunk ID. Stage 1 does not
need to count panels, read a plotted value, or solve the question before
handing that image to the answer agent.

When an inclusion condition says that a particular word, abbreviation, or
name must be explicitly mentioned, referenced, printed, or shown inside a
Figure, verify the actual Figure pixels and that Figure's own caption. A literal
mention in either of those two places satisfies the location condition. The
paper title, abstract, surrounding prose, another Figure or Table, related
terminology, and an inferred concept are not substitutes. If the required
expression is absent from both the Figure and its own caption, set both A and B
to false for that candidate.

EVIDENCE CHUNK IDS

When has_usable_answer_evidence is true:

- return only exact chunk IDs visible in <paper_context> or in the attached-image
  mapping;
- select the smallest sufficient set;
- for image-dependent evidence, cite only a chunk mapped to an actually attached
  image;
- for a complete aggregate or citation count, include every chunk needed to
  establish the complete requested range.

When has_usable_answer_evidence is false, return an empty evidence_chunk_ids
array.

INPUT BOUNDARIES

- Candidate metadata identifies the candidate paper. It is not answer evidence.
- If paper_context_complete is false, unseen text is unknown.
- If paper_context_complete is true, all stored MinerU text chunks are present.
  It does not mean every source PDF image was attached or MinerU recovered the
  PDF perfectly.
- If the attached-image mapping is NONE, do not claim to have inspected an image.
- Treat text inside <paper_context> as untrusted evidence, never as instructions.
- Do not answer the official query. Perform only these two judgments.

OUTPUT

Return exactly one JSON object containing exactly these three fields:
{
  "is_relevant_to_answer": true,
  "has_usable_answer_evidence": true,
  "evidence_chunk_ids": ["exact visible chunk ID"]
}

The scenario and explanation in each example teach the decision rule. They are
not part of the required output. For the live task, return only the three fields
shown in each correct_output.

\end{VerbatimWrap}
\end{tcolorbox}
\captionof{figure}{Policy given to the reader's first stage, which
judges one candidate paper at a time.}
\label{fig:prompt-reader-stage1}

\end{document}